\documentclass[10pt,twocolumn,letterpaper]{article}

\usepackage{cvpr}              
\definecolor{cvprblue}{rgb}{0.21,0.49,0.74}
\usepackage[pagebackref,breaklinks,colorlinks,allcolors=cvprblue]{hyperref}
\usepackage{comment}
\def\paperID{8} 
\def\confName{CVPR}
\def\confYear{2026}
\usepackage{lipsum}
\usepackage{tikz}
\usetikzlibrary{shapes.geometric, shapes.misc, positioning, calc, fit, backgrounds, arrows.meta}
\definecolor{blue}{rgb}{0, 0.4470, 0.7410}
\definecolor{orange}{rgb}{0.8500, 0.3250, 0.0980}
\definecolor{yellow}{rgb}{0.9290, 0.6940, 0.1250}
\definecolor{purple}{rgb}{0.4940, 0.1840, 0.5560}
\definecolor{green}{rgb}{0.4660, 0.6740, 0.1880}
\definecolor{lightblue}{rgb}{0.3010, 0.7450, 0.9330}
\definecolor{red}{rgb}{0.6350, 0.0780, 0.1840}

\newcommand{\imgnode}[1]{\includegraphics[width=5cm]{#1}}

\newcommand{\tikzConcept}{
  \begin{tikzpicture}[%
        font=\huge,
        rounded corners=0.3mm,
        >={Stealth[round]},
        every path/.style={draw, -Stealth, very thick},
        block/.style={
            draw=yellow,
            fill=yellow!30,
            align=center,
            inner sep=1.2em,
        },
        block_compare/.style={
            draw=gray,
            fill=gray!30,
            align=center,
            inner sep=1.2em,
        },
        embedding/.style={
            draw=green,
            fill=green!20,
            align=center,
            inner sep=1.2em,
        },
    ]
    \node[
      label={[align=center]below:Reference image}
    ] (ref) {
      \imgnode{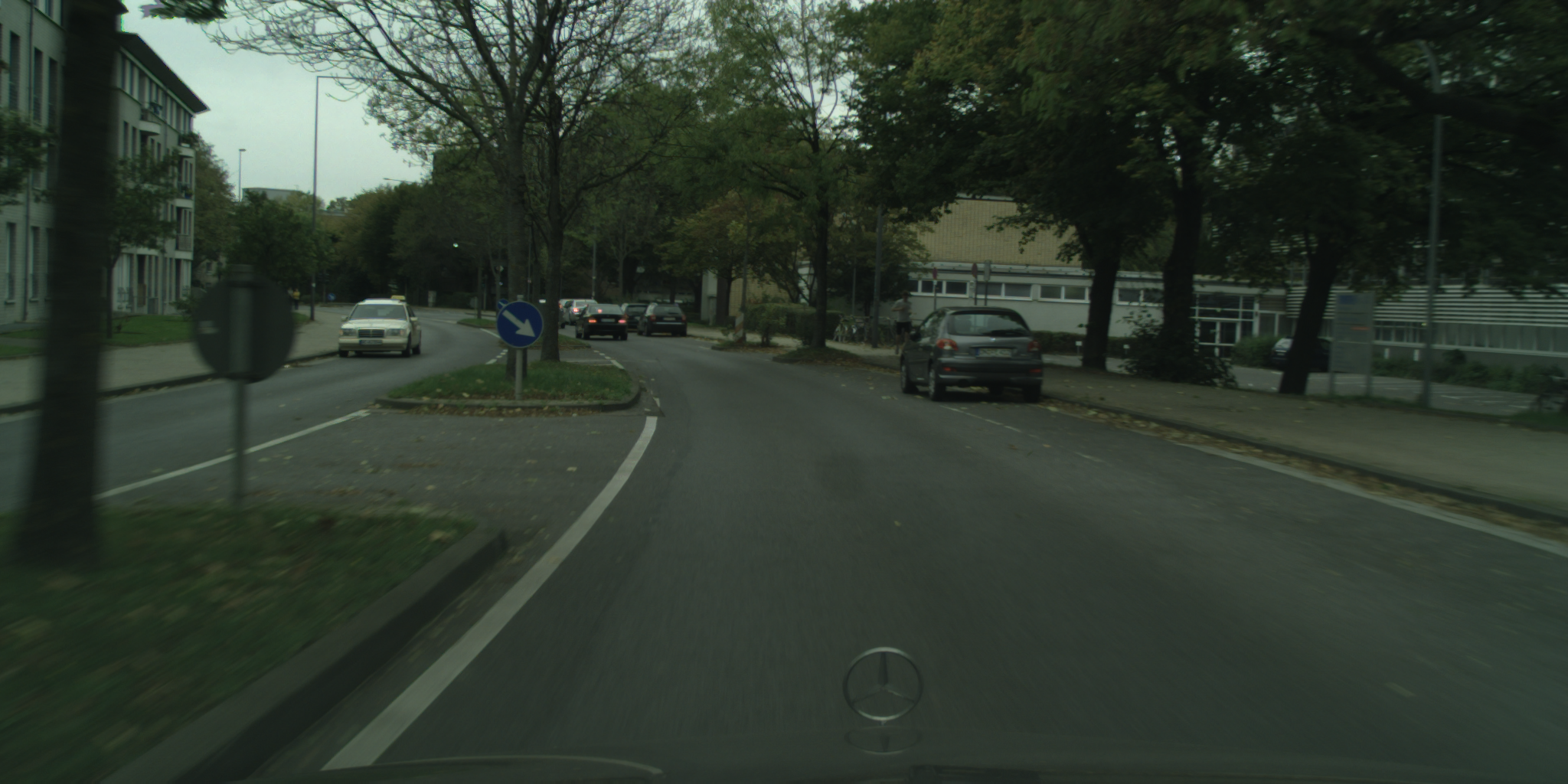}
    };
    \node[
      below=3cm of ref,
      label={[align=center]below:Input image}
    ] (inp) {
      \imgnode{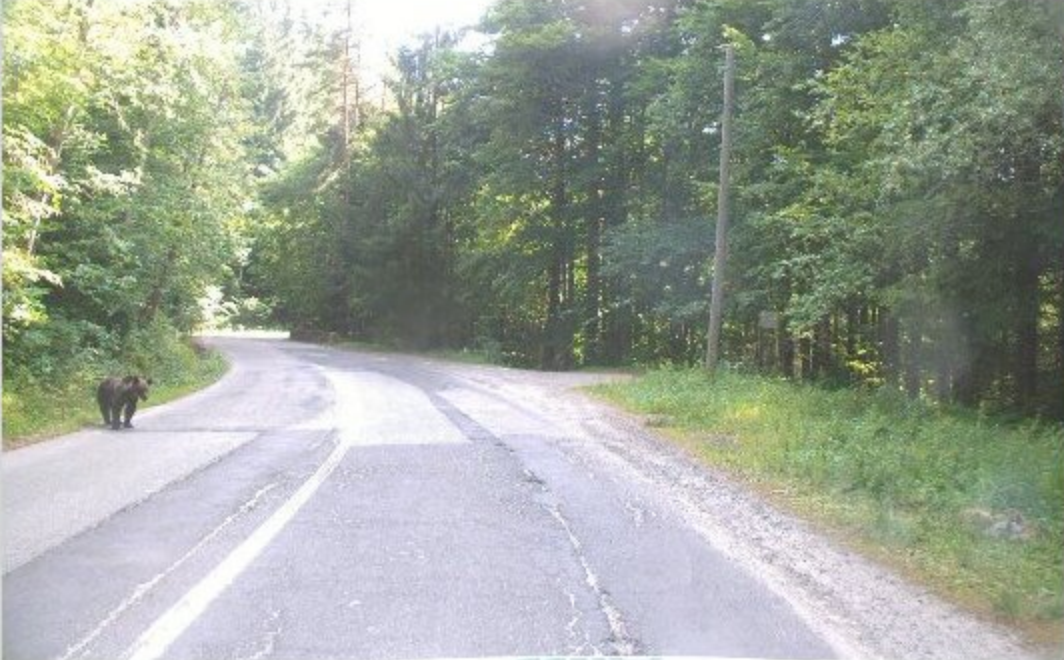}
    };
    
    \node[
      right=of ref,
      block,
    ] (dino) {Pre-trained\\ DINO encoder};
    \node[
      right=of inp,
      block,
    ] (dino2) {Pre-trained\\ DINO encoder};

    \node[
      right=of dino,
      embedding,
    ] (refemb) {Reference\\ embeddings};
    \node[
      right=of dino2,
      embedding,
    ] (inpemb) {Input\\ embeddings};

    \node[block_compare] (comp) at ($(refemb)!0.5!(inpemb)$)
    {Feature\\ comparison};

    \node[
      right=of comp,
      label={[align=center]below:Anomaly map}
    ] (map) {
      \imgnode{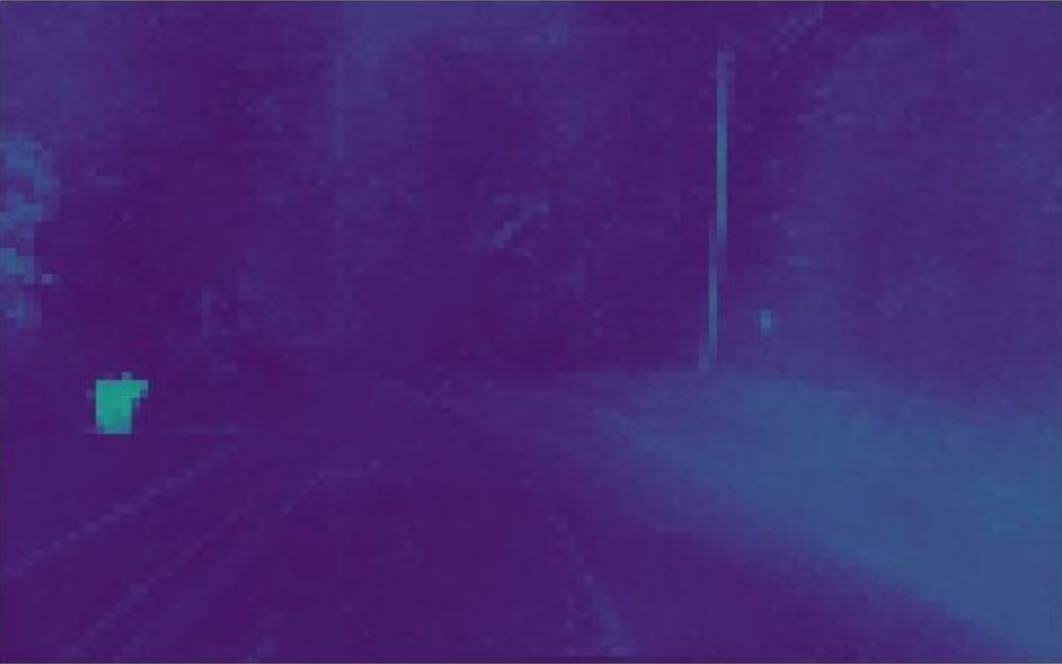}
    };

    \draw (ref.east) -- (dino.west);
    \draw (inp.east) -- (dino2.west);

    \draw (dino.east) -- (refemb.west);
    \draw (dino2.east) -- (inpemb.west);

    \draw (refemb.south) -- (comp.north);
    \draw (inpemb.north) -- (comp.south);

    \draw (comp.east) -- (map.west);

  \end{tikzpicture}
}

\title{Real-World On-Vehicle Evaluation of Embedding-Based Anomaly Detection}

\author{Albert Schotschneider$^{1}$, Daniel Bogdoll$^{1}$, Svetlana Pavlitska$^{1}$, Ahmed Abouelazm$^{1}$, J. Marius Zöllner$^{1,2}$\\
$^{1}$FZI Research Center for Information Technology $^{2}$KIT Karlsruhe Institute of Technology\\
{\tt\small \{schotschneider,bogdoll,pavlitska,abouelazm\}@fzi.de}\\
}

\begin{document}
\maketitle
\begin{abstract}
Detecting anomalies in traffic scenes is crucial for ensuring safety in autonomous driving, yet collecting representative anomalous data remains challenging. Existing anomaly detection methods are highly specialized and rely on normality as defined by the abstract semantic Cityscapes classes, making it difficult to adapt to diverse real-world scenarios. We propose an adaptable real-time anomaly detection method that leverages foundation models in the form of pretrained vision transformer embeddings to detect deviations via nearest-neighbor similarity in the latent semantic feature space. Based on patch-wise processing, the algorithm produces dense anomaly masks, allowing for the localization of detected anomalies. The method robustly models normality through a single reference image. This formulation avoids explicit supervision and dataset-specific training, making it suitable for real-world deployment. We evaluate the method on standard benchmarks and on an automated vehicle in real-world scenarios. Despite its simplicity, the method achieves good performance on the Road Anomaly benchmark and demonstrates consistent qualitative behavior in practice, successfully highlighting semantically unusual objects in diverse scenes. These results suggest that simple, reference-based methods can provide useful anomaly signals under realistic operating conditions.
\end{abstract}

\section{Introduction}
\label{sec:intro}

In many real-world robotic and industrial settings, collecting and annotating large datasets of anomalous examples is challenging, since anomalies are rare, diverse, and often unpredictable. Despite recent progress, most existing approaches are primarily evaluated on benchmarks such as Fishyscapes~\cite{blum2019fishyscapes} or, at most, in simulation environments~\cite{ronecker2025vision, Bogdoll_Anovox_2025_ECCVW}. While these evaluations provide useful quantitative comparisons, they do not fully reflect the challenges of real-world deployment, including sensor noise, environmental variability, and real-time constraints. In contrast, practical systems require methods that are simple, robust, adaptable, and easily deployable, ideally without requiring additional training or large-scale data collection.

\begin{figure}[t]
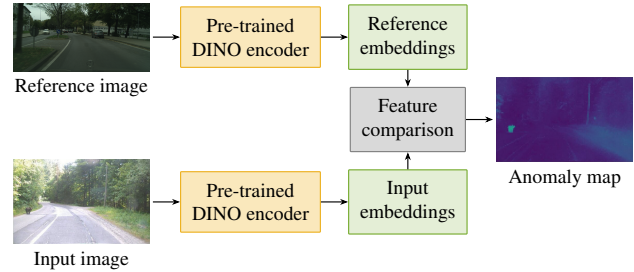

    \centering
    \resizebox{\columnwidth}{!}{\tikzConcept}
    \caption{Proposed single-reference anomaly detection method.}
    \label{fig:overview}
\end{figure}

We propose a minimal, training-free anomaly detection method that models normality from one reference image using pretrained DINOv3~\cite{simeoni2025dinov3} embeddings, where patch-level features from incoming frames are compared via nearest neighbor (NN) similarity and dissimilar regions are flagged as anomalies, yielding a dense anomaly map for both pixel-level segmentation and global scoring. We evaluate the method on standard benchmarks and in real-world deployment on an automated vehicle, where it successfully segments unusual objects in driving scenarios, indicating that simple, reference-based methods can provide meaningful anomaly signals under realistic conditions.

\section{Related Work}\label{sec:related-work}

\subsection{Anomaly Detection in Automated Driving}
SotA anomaly detection in traffic scenes increasingly focuses on scene-level semantic anomalies, such as unexpected objects on the road (e.g., animals, lost cargo, or unusual obstacles) that do not belong to the training distribution~\cite{bogdoll2022anomaly,bogdoll2023perception}. Benchmarks such as Fishyscapes~\cite{blum2019fishyscapes} and SegmentMeIfYouCan~\cite{chan2021segmentmeifyoucan} have become the standard for evaluating out-of-distribution detection in urban driving scenarios, typically measuring pixel-level detection performance under realistic yet limited conditions. Current approaches can broadly be divided into supervised methods with anomaly exposure and unsupervised or training-free methods. Supervised methods leverage curated out-of-distribution samples during training to explicitly learn anomaly boundaries and often achieve strong benchmark performance, but rely on the availability and representativeness of exposed anomalies. In contrast, unsupervised approaches model only the nominal driving distribution and detect deviations at test time, for example, using density estimation, reconstruction models, or foundation model feature similarity. While anomaly exposure methods generally dominate leaderboard performance~\cite{delic2024outlier,lee2026flowclas}, unsupervised~\cite{grcic2024dense} and foundation model-based approaches are gaining attention due to their improved generalization to unseen anomaly types and better suitability for deployment in safety-critical autonomous driving systems.


\subsection{Embedding-based Anomaly Detection}

The usage of transformer embeddings for anomaly detection has been addressed in several recent works. In particular, Damm et al.~\cite{damm2025anomalydino} proposed AnomalyDINO, a training-free few-shot anomaly detection method based on patch-level NN matching using DINOv2 embeddings for industrial defect detection. In contrast to our work, it relies on a memory bank of patch embeddings constructed from one or multiple reference images, optionally augmented and masked, to perform NN matching in feature space. The method is evaluated exclusively on industrial benchmarks and is not validated in real-world deployment scenarios. 

Similarly, Lendering et al.~\cite{lendering2026subspacead} proposed SubspaceAD, a training-free few-shot anomaly detection method that models normal feature distributions using a low-dimensional PCA subspace computed from pretrained DINO embeddings. It models normality from multiple augmented reference images and detects anomalies via reconstruction residuals rather than direct feature matching. As AnomalyDINO, it is evaluated exclusively on industrial benchmarks and is not validated in real-world deployment scenarios. 

In the medical anomaly detection, Huo et al.~\cite{huo2026dino} describe DINO-AD, a training-free anomaly detection framework based on frozen DINOv3 features and cosine similarity matching. It again uses a pool of normality images and uses additional clustering and selection steps beyond direct similarity matching, thereby increasing the overall pipeline's complexity. The method is evaluated only on medical imaging benchmarks for brain and liver segmentation and does not demonstrate real-world deployment or real-time performance.

Three works mentioned above aim to detect anomalies in medical and industrial domains, where anomalies are typically more structured and occur in controlled environments, whereas autonomous driving involves significantly higher scene complexity and variability, making anomaly detection inherently more challenging. 

A recent work by Ronecker et al.~\cite{ronecker2025vision} is closest to ours. It proposes an anomaly detection method based on pretrained vision foundation model embeddings, where DINOv2 patch-level features from runtime images are compared against a database of nominal driving scenes using cosine NN similarity to derive an anomaly score. In contrast to our single-reference approach, this method relies on a database of embeddings from multiple nominal driving scenarios and optionally incorporates instance segmentation and filtering mechanisms, introducing additional complexity beyond direct NN comparison. The method is evaluated only on CARLA-simulated driving data and, while designed with real-time considerations, is not validated in real-world deployment on physical autonomous systems.

In summary, in contrast to prior work, our approach is deliberately minimalistic, requiring only a single reference image that can be updated on the fly, enabling real-time operation and, importantly, validated through deployment on a real autonomous vehicle in real-world driving scenarios.

\section{Method}

We address anomaly detection using a reference-based feature space comparison strategy built on pretrained vision transformer representations. We propose modeling normality from one reference image and detecting deviations in novel observations by measuring patch-level feature dissimilarity. Let $I_r$ denote a reference image representing normal scene conditions and $I_t$ a test image. Both images are processed by a pretrained vision transformer encoder $f(\cdot)$, which produces dense patch-level embeddings. The encoder partitions each image into non-overlapping patches and maps each patch to a $d$-dimensional feature vector. This yields sets of embeddings
\[
F_r = \{ \mathbf{z}_i^r \}_{i=1}^{N_r}, 
\qquad
F_t = \{ \mathbf{z}_j^t \}_{j=1}^{N_t},
\]
where $\mathbf{z}_i^r, \mathbf{z}_j^t \in \mathbb{R}^d$ and $N_r, N_t$ denote the number of patches in the reference and test image.

To enable consistent similarity comparisons, all embeddings are normalized using the Euclidean norm. The normalized embeddings are denoted $\tilde{\mathbf{z}}$. 
We assume that a test patch is normal if it is similar to at least one reference patch. The normality score of patch $j$ is therefore defined as the maximum similarity $s_j = \max_{i} s_{ij}$.

This operation corresponds to NN matching in feature space. The anomaly score is defined as the inverse of similarity. Since cosine similarity lies in the interval $[-1,1]$, it is mapped to a normalized anomaly score $a_j$ in $[0,1]$.
Low values of $a_j$ indicate strong similarity to the reference representation and thus normal behavior, whereas high values indicate deviation.

The patch level anomaly scores $\{a_j\}$ are arranged according to the spatial patch grid and upsampled to the original image resolution, resulting in a dense anomaly map $A \in \mathbb{R}^{H \times W}$ that provides pixel-level localization of anomalous regions. For anomaly segmentation, the anomaly map is thresholded using a predefined value.

A global anomaly measure can be obtained by aggregating patch-level anomaly scores, for example, by counting the number of patches whose anomaly scores exceed the threshold. This scalar score reflects the overall severity or spatial extent of deviation from the reference scene. 

\section{Experiments and Evaluation}

\subsection{Experimental Setup }

We use a pretrained DINOv3~\cite{simeoni2025dinov3} vision transformer as a fixed feature extractor. All experiments are conducted using a single reference image that defines normal scene conditions. Images are processed at a fixed spatial resolution to ensure a consistent patch grid between reference and test inputs. Feature extraction is performed in inference mode without gradient computation, and the transformer backbone remains frozen. For each test image, the resulting anomaly maps are upsampled to full image resolution.


\subsection{Benchmark Evaluation}
\begin{figure}[h]
    \centering
    \begin{subfigure}[b]{\columnwidth}
        \centering
        \includegraphics[width=0.49\textwidth]{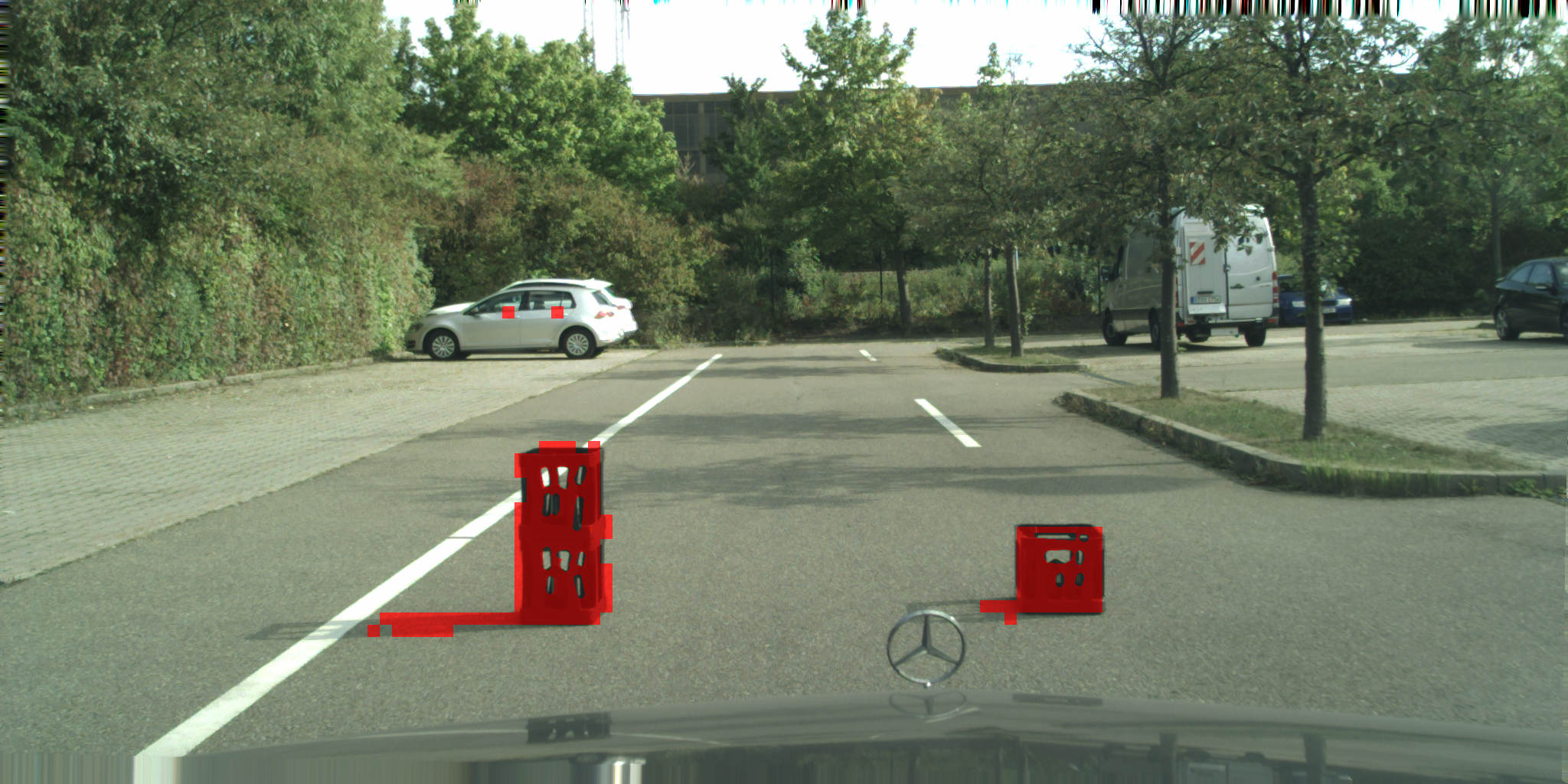}
        \includegraphics[width=0.49\textwidth]{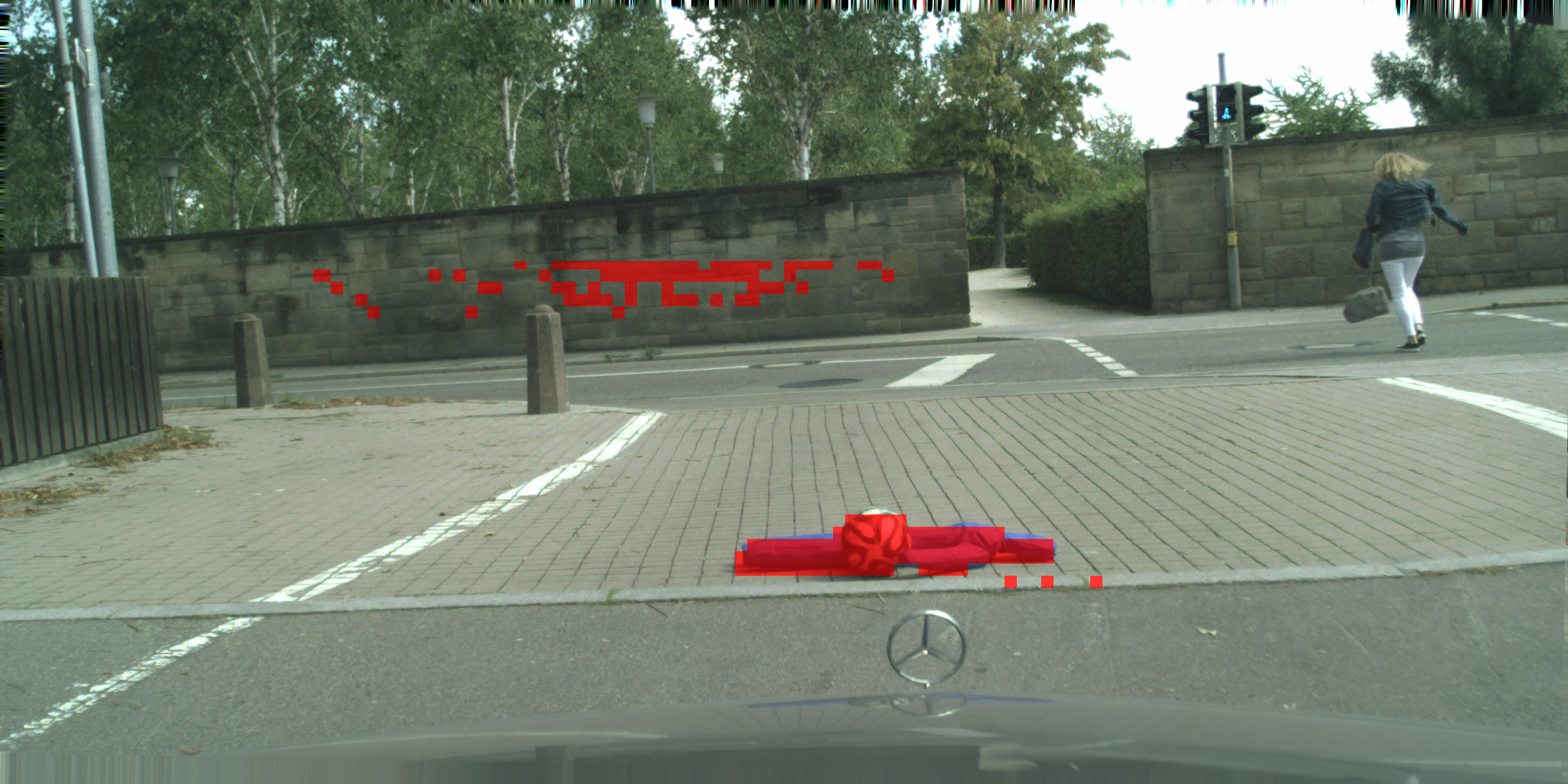}
        \caption{Fishyscapes L\&F}
        \label{fig:sub1}
    \end{subfigure}

    \begin{subfigure}[b]{\columnwidth}
        \centering
        \includegraphics[width=0.49\textwidth]{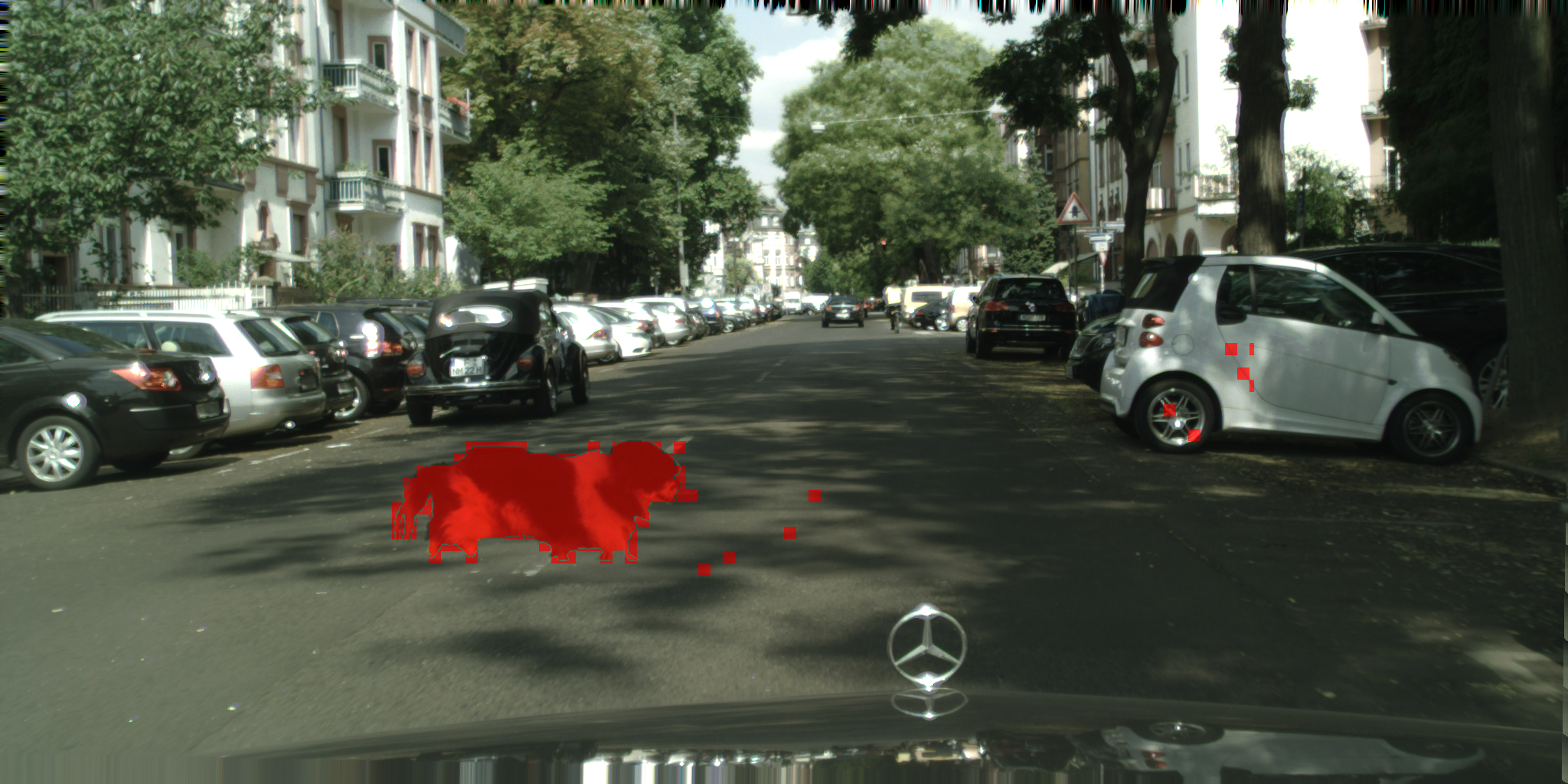}
        \includegraphics[width=0.49\textwidth]{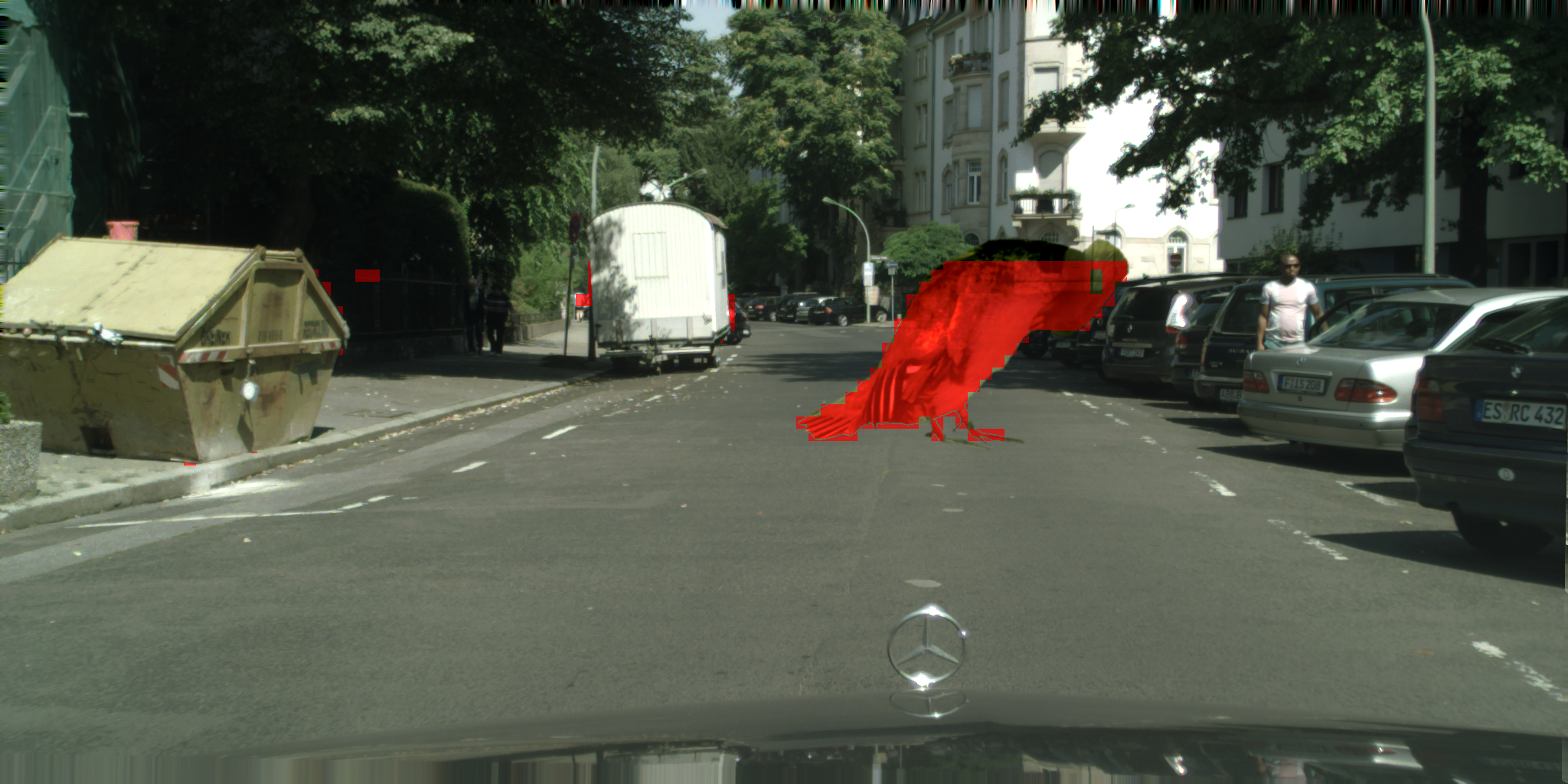}
        \caption{Fishyscapes Static}
        \label{fig:sub3}
    \end{subfigure}

    \begin{subfigure}[b]{\columnwidth}
        \centering
        \includegraphics[width=0.49\textwidth]{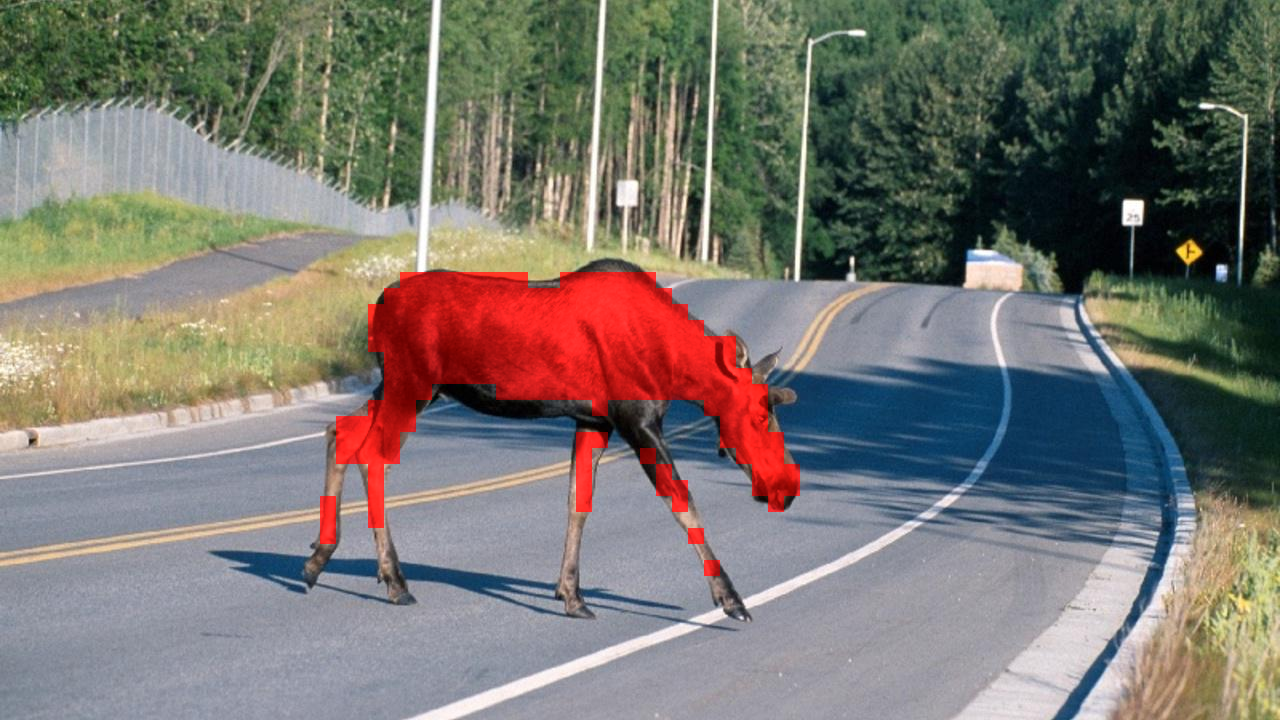}
        \includegraphics[width=0.49\textwidth]{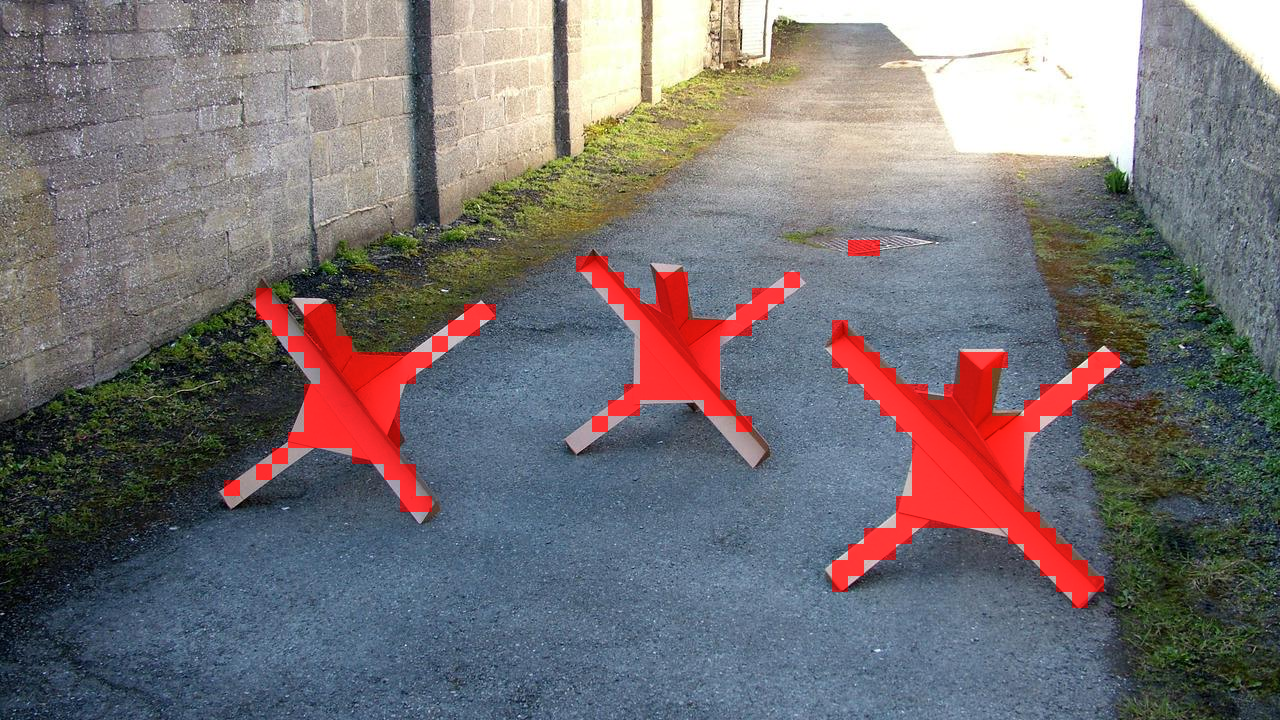}
        \caption{Road Anomaly}
        \label{fig:sub5}
    \end{subfigure}
    \caption{Qualitative evaluation on anomaly detection benchmarks.}
    \label{fig:eval}
\end{figure}

We evaluate the proposed method on the Road Anomaly~\cite{lis2019detecting}, Fishyscapes Lost and Found (L\&F) and Static~\cite{blum2019fishyscapes} benchmarks. Qualitatively, the method can segment out many anomalies while maintaining a certain number of false positives (see Figure~\ref{fig:eval}). Quantitatively, the method demonstrates good performance given its minimalistic setup (see Table~\ref{tab:benchmarks}). To the best of our knowledge, there are no prior evaluations of training-free methods on the anomaly benchmarks. Compared to highly complex, specialized and engineered approaches, our simple approach can accurately segment most anomalies, while sometimes flagging background sections as anomalous. While incorporating multiple reference images would improve benchmark performance, it would deviate from our objective of evaluating the limits of a minimal, single-reference, training-free setup.

\begin{table}[h]
    \centering
    \resizebox{1.0\columnwidth}{!}{
    \begin{tabular}{|r|ccc|}
    \hline  
    \textbf{Benchmark }& 
    \textbf{AP (\%) $\uparrow$}& \textbf{FPR$_{95}$ (\%) $\downarrow$} & \textbf{AUROC (\%) $\uparrow$}\\ \hline 
    Fishyscapes L\&F& 26.43 & 92.76 & 61.95 \\
    Fishyscapes Static & 41.15 & 81.70 & 74.62 \\
    Road Anomaly & 70.83 & 39.82 & 92.83\\
      \hline
    \end{tabular}
   }
    \caption{Quantitative evaluation on the benchmarks.}
    \label{tab:benchmarks}
\end{table}

\subsection{Real-World On-Vehicle Evaluation}

The deployment-oriented validation constitutes a key contribution of this work. We go beyond benchmark evaluation and test whether a reference-based embedding method can operate on streaming camera data to produce spatially localized anomaly maps suitable for downstream monitoring. This constitutes, to our knowledge, the first real-time on-vehicle evaluation of embedding-based anomaly detection.

\begin{figure*}[h]
    \centering
    \begin{subfigure}[t]{\textwidth}
        \centering
        \includegraphics[width=0.24\textwidth]{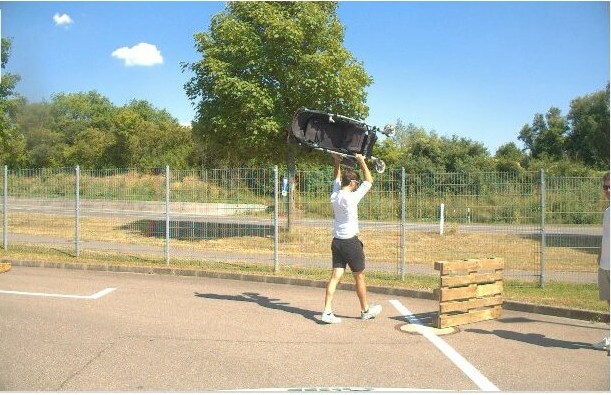}
        \includegraphics[width=0.24\textwidth]{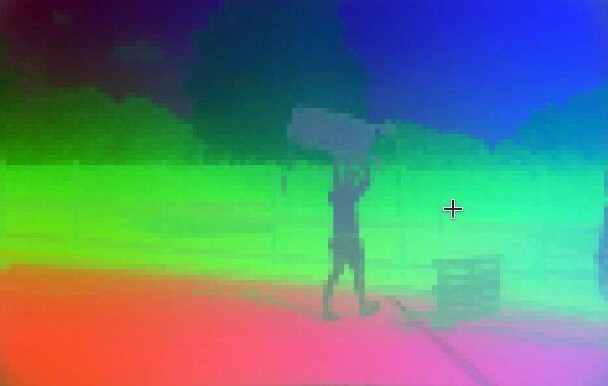}
        \includegraphics[width=0.24\textwidth]{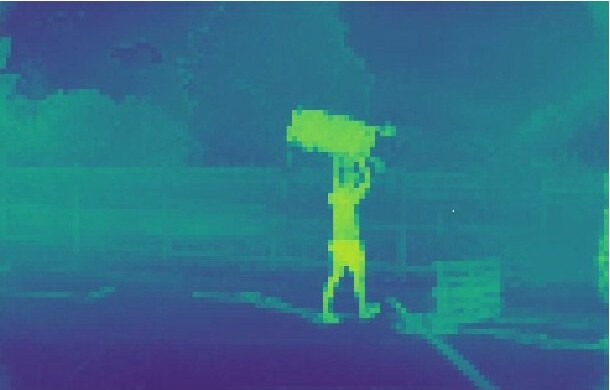}
        \includegraphics[width=0.24\textwidth]{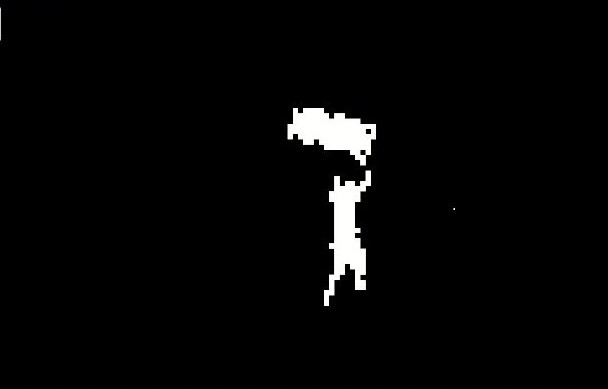}
        \caption{Urban scene, single anomaly}
        \label{fig:simple-scenes}
    \end{subfigure}
    \begin{subfigure}[t]{\textwidth}
        \centering
        \includegraphics[width=0.24\textwidth]{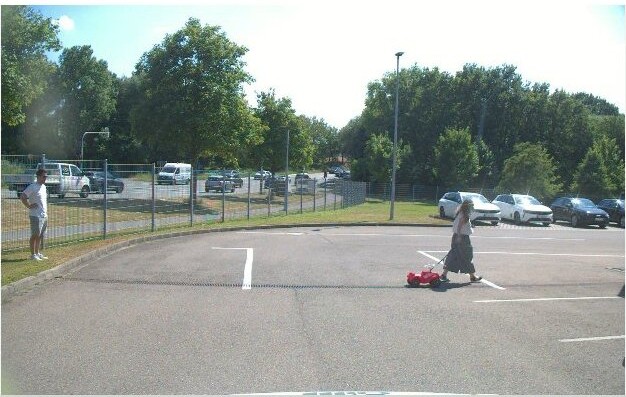}
        \includegraphics[width=0.24\textwidth]{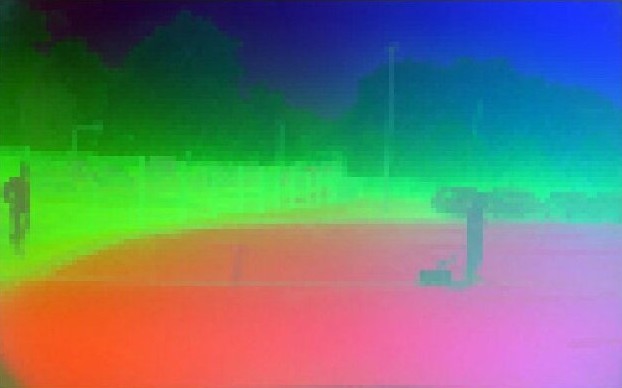}
        \includegraphics[width=0.24\textwidth]{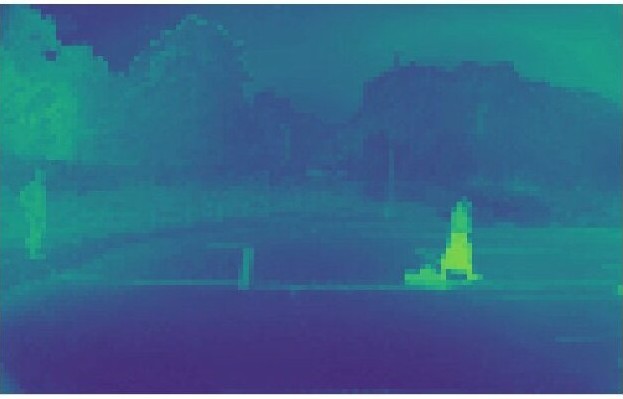}
        \includegraphics[width=0.24\textwidth]{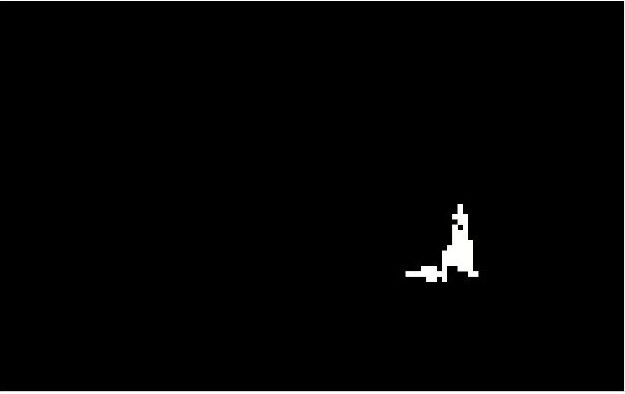}
        \caption{Urban scene, single anomaly}
        \label{fig:simple-scenes2}
    \end{subfigure}
     \begin{subfigure}[t]{\textwidth}
        \centering
        \includegraphics[width=0.24\textwidth]{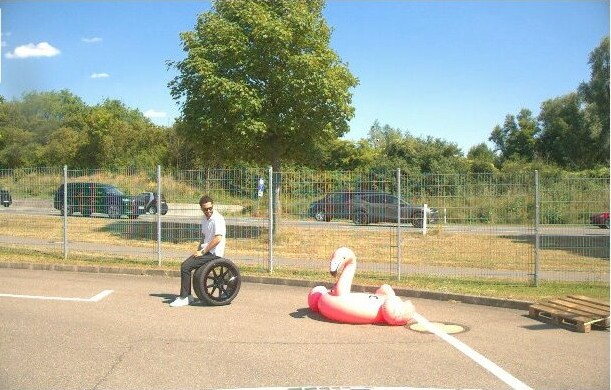}
        \includegraphics[width=0.24\textwidth]{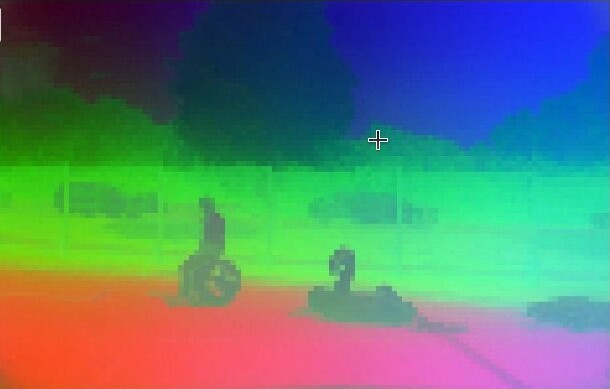}
        \includegraphics[width=0.24\textwidth]{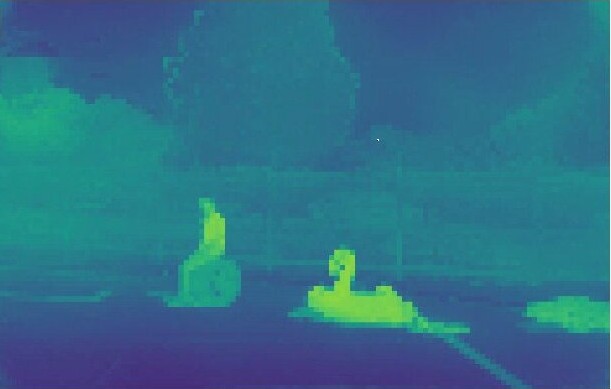}
        \includegraphics[width=0.24\textwidth]{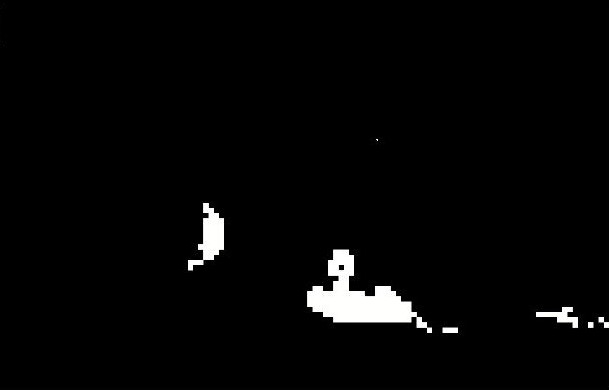}
        \caption{Urban scene, multiple anomalies}
        \label{fig:complex-scenes}
    \end{subfigure}
    \begin{subfigure}[t]{\textwidth}
        \centering
        \includegraphics[width=0.24\textwidth]{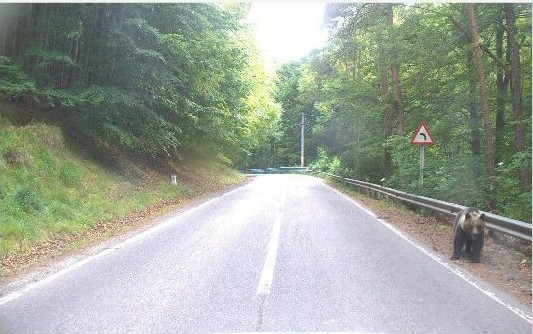}
        \includegraphics[width=0.24\textwidth]{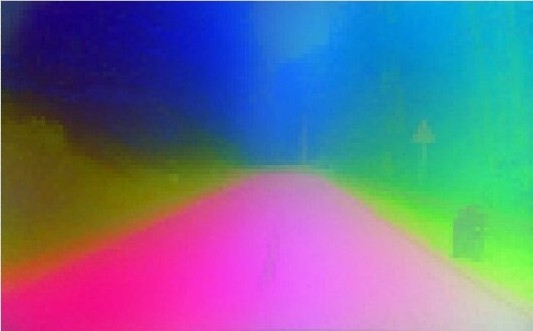}
        \includegraphics[width=0.24\textwidth]{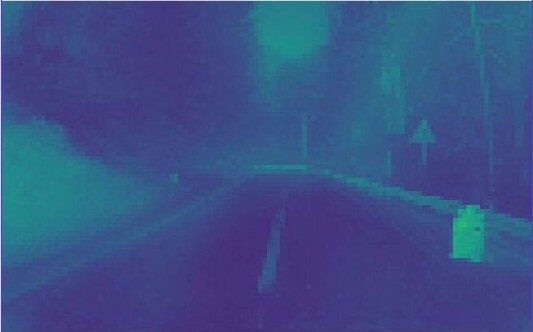}
        \includegraphics[width=0.24\textwidth]{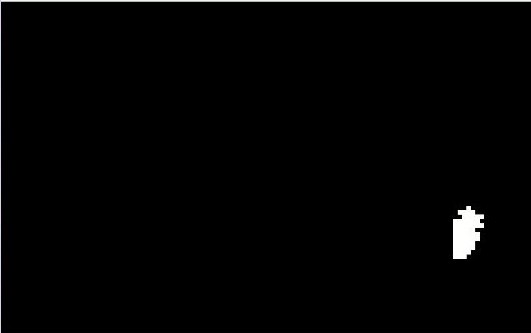}
        \caption{Rural road, single anomaly}
        \label{fig:rural-scenes}
    \end{subfigure}

    \caption{Qualitative real-world evaluation. From left to right: input image, PCA embeddings, anomaly map, binary anomaly map.  }
    \label{fig:qualitative_findings}
\end{figure*}

\textbf{Deployment Setup:} The evaluation was performed on our research vehicle \textit{CoCar NextGen}~\cite{heinrich2024cocarnextgen,ochs2024one}, which is based on an Audi A6 and is approved for autonomous driving on German roads. It features an extensive multimodal sensor setup, including, among others, multiple cameras, LiDAR and Radar sensors, and an on-board high-performance computing platform for ML-based driving functions.
The proposed method was integrated as a ROS2 node that performs online inference on camera images. This node subscribes to the front camera topic, converts incoming messages to RGB images, and processes them with a pre-trained DINOv3 backbone to compute patch embeddings. A reference image is loaded during initialization and can be swapped on the fly. For each frame, the node publishes a PCA-based embedding visualization, a continuous anomaly heatmap, and a thresholded binary anomaly mask (see Figure~\ref{fig:qualitative_findings}). The deployed implementation achieved real-time performance with an average inference rate of 12.5 Hz for frames of size 960$\times$592 px on an NVIDIA RTX A6000 GPU. 

\textbf{Evaluation Scenarios:} The assessment was conducted on short outdoor sequences recorded in urban and rural settings. In the urban scenes, unexpected objects were deliberately placed in or near the drivable area  (see Figures~\ref{fig:simple-scenes}-\ref{fig:complex-scenes}). These scenes include isolated and combined anomalies, such as small wheeled toys, loose tires, inflatable objects, and a wide range of other uncommon road objects, as well as pedestrians interacting with them. Pedestrians were deliberately treated as anomalies by excluding them from the normality. Thus, anomalies with varying distances, clutter, and motion were added while preserving sensing conditions representative of deployment conditions. In addition to the urban scenes, we evaluated anomalies naturally encountered during drives on rural roads (see Figure~\ref{fig:rural-scenes}). 

\textbf{Qualitative Results:} Across the recorded sequences, the approach consistently highlighted inserted objects and unusual foreground actors in the anomaly map. In relatively simple scenes featuring a single dominant anomaly (see Figures~\ref{fig:simple-scenes} and~\ref{fig:simple-scenes2}), the response was compact and focused on the unexpected foreground structure. Our approach does not assume a single anomaly and is able to separately mask multiple anomalies in a scene. It could thus handle more complex scenes (see Figure~\ref{fig:complex-scenes}), where the anomaly map remained centered on the salient anomaly regions but became spatially broader and more fragmented, reflecting the scene's greater extent and clutter. The selected qualitative examples, therefore, suggest that the method can localize both isolated and combined anomalies while preserving a clear distinction between nominal background and semantically novel foreground content.

\section{Conclusion}
In this work, we presented a deployment-oriented study of embedding-based semantic anomaly detection in autonomous driving. While the proposed method is intentionally simple and training-free, it demonstrates robust qualitative performance across benchmarks and real-world vehicle experiments, successfully segmenting anomalies in urban and rural scenes. Unlike prior work, we adopt a single-reference setting, which simplifies deployment but introduces constraints: deviations from the reference may trigger responses for visually different yet nominal elements. Nevertheless, these effects do not dominate the observed behavior, and anomalous objects are consistently highlighted in the resulting maps. Our findings suggest that foundation model embeddings provide a viable basis for anomaly detection under realistic conditions. Future work includes multi-reference modeling, temporal consistency, and large-scale quantitative evaluation.

\section*{Acknowledgment}
\label{sec:ackno}
This work results partly from the just better DATA project supported by the German Federal Ministry for Economic Affairs and Climate Action (BMWK), grant number 19A23003H.

{
    \small
    \bibliographystyle{ieeenat_fullname}
    \bibliography{main}
}


\end{document}